\documentclass[letterpaper]{article} 
\usepackage{aaai25}  
\usepackage{times}  
\usepackage{helvet}  
\usepackage{courier}  
\usepackage[hyphens]{url}  
\usepackage{graphicx} 
\usepackage{natbib}  
\usepackage{caption} 
\usepackage{xspace}
\usepackage{amsmath}
\usepackage{amsfonts}
\usepackage{multirow}
\usepackage{booktabs}
\usepackage[linesnumbered,ruled]{algorithm2e}

\renewcommand{\vec}[1]{\mathbf{#1}}

\newcommand{\ns}{(\vec{x}^{[i]}, \hat{\vec{y}}^{[i]})}
\newcommand{\eg}{\emph{e.g.,}\xspace}

\newcommand{\ie}{\emph{i.e.,}\xspace}

\newcommand{\quotes}[1]{``#1''}

\title{Enhanced Sample Selection with Confidence Tracking: Identifying Correctly Labeled yet Hard-to-Learn Samples in Noisy Data}
\author{
    Weiran Pan\textsuperscript{\rm 1, \rm 2},
    Wei Wei\textsuperscript{\rm 1, \rm 2}\thanks{Corresponding author.},
    Feida Zhu\textsuperscript{\rm 3},
    Yong Deng\textsuperscript{\rm 4}
}
\affiliations{
    \textsuperscript{\rm 1} School of Computer Science and Technology, Huazhong University of Science and Technology, China\\
    \textsuperscript{\rm 2} Joint Laboratory of HUST and Pingan Property \& Casualty Research (HPL), China\\
    \textsuperscript{\rm 3} School of Computing and Information Systems, Singapore Management University, Singapore\\
    \textsuperscript{\rm 4} State Grid Fujian Electric Power Co.\\
    \{panwr, weiw\}@hust.edu.cn, fdzhu@smu.edu.sg, hhdyyong@qq.com
}

\begin{document}

\maketitle

\begin{abstract}
We propose a novel sample selection method for image classification in the presence of noisy labels. Existing methods typically consider small-loss samples as correctly labeled. However, some correctly labeled samples are inherently difficult for the model to learn and can exhibit high loss similar to mislabeled samples in the early stages of training. Consequently, setting a threshold on per-sample loss to select correct labels results in a trade-off between precision and recall in sample selection: a lower threshold may miss many correctly labeled hard-to-learn samples (low recall), while a higher threshold may include many mislabeled samples (low precision). To address this issue, our goal is to accurately distinguish correctly labeled yet hard-to-learn samples from mislabeled ones, thus alleviating the trade-off dilemma. We achieve this by considering the trends in model prediction confidence rather than relying solely on loss values. Empirical observations show that only for correctly labeled samples, the model's prediction confidence for the annotated labels typically increases faster than for any other classes. Based on this insight, we propose tracking the confidence gaps between the annotated labels and other classes during training and evaluating their trends using the Mann-Kendall Test. A sample is considered potentially correctly labeled if all its confidence gaps tend to increase. Our method functions as a plug-and-play component that can be seamlessly integrated into existing sample selection techniques. Experiments on several standard benchmarks and real-world datasets demonstrate that our method enhances the performance of existing methods for learning with noisy labels.
\end{abstract}

\begin{links}
    \link{Code}{https://github.com/Aliinton/ConfidenceTracking}
\end{links}

\section{Introduction}
The remarkable success of deep learning methods in classification tasks can largely be attributed to high-quality datasets. However, collecting such datasets through manual labeling can be both time-consuming and expensive in many applications. Acquiring data via online queries~\cite{webvision} or crowdsourcing~\cite{clothing1m} can construct large-scale datasets at a lower cost but inevitably introduces noise labels. Existing research~\cite{DBLP:conf/iclr/ZhangBHRV17} has shown that deep neural networks can easily fit noisy data, resulting in poor generalization. Therefore, developing algorithms robust to noisy labels is of great practical importance~\cite{DBLP:conf/nips/NatarajanDRT13}.

\begin{figure*}
    \centering
    \includegraphics[width=0.95\linewidth]{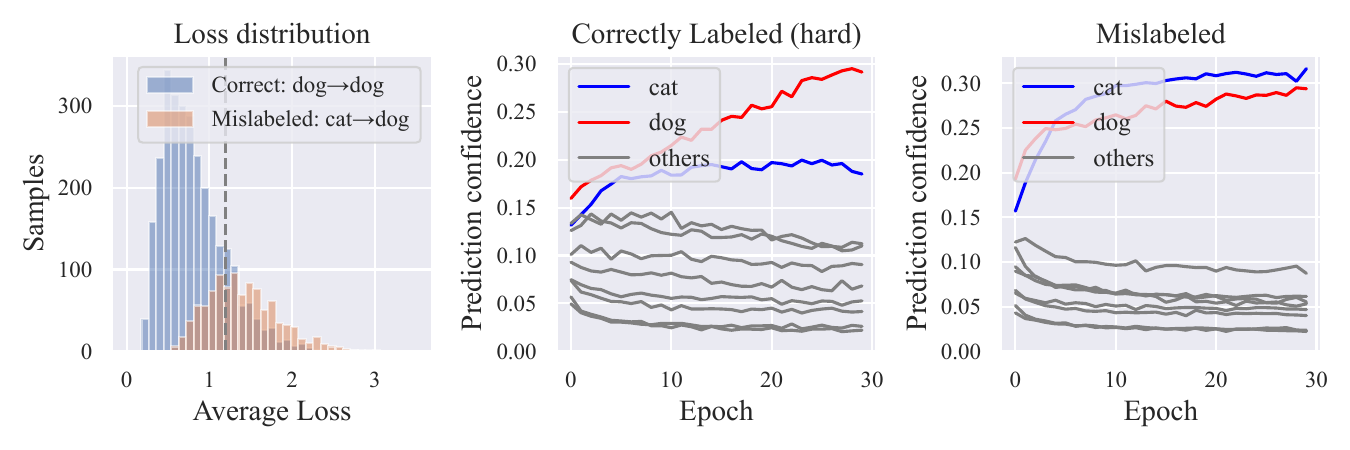}
    \caption{Illustration of Confidence Tracking. We train a PreActResNet-18 model using cross-entropy loss and an SGD optimizer on CIFAR-10N-Worst (CIFAR-10 dataset with human-annotated real-world noisy labels, its noise rate is 40.21\%). 
    The left graph presents the average per-sample loss distribution in the first 30 epochs. We regard samples with an average loss greater than 1.2 (indicated by the dotted vertical line) as hard-to-learn ones and show the model prediction confidence trajectories on hard-to-learn dogs' images (middle graph) and mislabeled cats' images (right graph).
    }
    \label{fig:trainingdynamics}
\end{figure*}

Sample selection methods aim to identify correct labels from noisy data, which is increasingly crucial for current deep learning models to effectively learn from noisy labels. There is a general consensus that the small-loss criterion is an effective approach, which assumes samples with small losses are more likely to have correct labels. In this context, the Co-teaching family~\cite{Co-teaching, Co-teaching+, jocor, CNLCU} and other state-of-the-art methods~\cite{DividMix, DISC} have been proposed. Typically, these methods select samples with losses below a threshold for training. So a higher threshold introduces more incorrect labels while a lower threshold excludes more correct labels, creating a trade-off dilemma between precision and recall in sample selection. To alleviate this issue, one possible solution is establishing another sample selection criterion that can distinguish correct labels from incorrect ones in high-loss data. Then we can combine it with the small-loss criterion to improve recall while maintaining precision in sample selection.

To achieve this, we propose considering the changing trends in model predictions to identify correct labels rather than relying solely on loss values. This is motivated by our empirical observations that although some correctly labeled samples may obtain similar loss values to mislabeled ones, their training dynamics are still distinguishable. As shown in Figure~\ref{fig:trainingdynamics}, some correctly labeled data can be hard to fit by the model and exhibit similar high loss to mislabeled data in the early training stage. It is difficult to distinguish them by setting a threshold on loss values. But we also observe that, only for the correctly labeled samples, the model's prediction confidence for annotated labels tends to rise more quickly than for other classes. For instance, with correctly labeled dog images, the model's prediction confidence for the \quotes{dog} class (\ie the posterior probability of the image belonging to the \quotes{dog} class predicted by the model) increases more rapidly than for any other class. Conversely, for cat images mislabeled as \quotes{dog}, the model's prediction confidence for the \quotes{cat} class rises slightly faster than for the \quotes{dog} class. This observation suggests it's possible to identify correctly labeled samples by considering training dynamics even when they are indistinguishable from mislabeled samples in terms of loss values.

Motivated by this finding, we propose a novel sample selection criterion that selects correct labels by monitoring how model predictions change during training, dubbed Confidence Tracking~(CT). Specifically, we track confidence gaps in model predictions between annotated labels and other classes during training. If all confidence gaps of a sample tend to increase, we regard it as a potentially correctly labeled sample. \emph{Unlike the small loss criterion which mainly considers the loss values, our method focuses on the changing trend in model predictions, allowing us to distinguish correctly labeled yet hard-to-learn samples from mislabeled ones within high-loss data.} In practice, our method functions as a plug-and-play component that can be combined with popular sample selection methods~\cite{M-correction, FINE, AUM, DISC} to enhance their performance. 

Our method is related to AUM~\cite{AUM} which also considers confidence gaps but selects samples with relatively large average logit margins as correctly labeled. Similar to the small-loss criterion, setting a threshold on average logit margins to select correct labels still faces the trade-off dilemma between precision and recall. Our experiments in Section~\ref{sec:main_results} demonstrate that CT accurately selects correct labels from samples rejected by AUM or the small-loss criterion, improving recall while maintaining precision in sample selection, and bringing performance gains to various benchmarks. To sum up, our key contributions are as follows:
\begin{itemize}
    \item We analyze why correctly labeled and mislabeled samples exhibit different training dynamics from the perspective of coherent gradients~\cite{coherentgradients} and provide supporting evidence.
    \item We propose a novel sample selection method based on monitoring changes in model predictions during training, termed Confidence Tracking~(CT), which is a plug-and-play component that can integrate with existing sample selection methods and accurately distinguish correctly labeled yet hard-to-learn samples from mislabeled ones.
    \item We experimentally show that our method improves the performance of the existing sample selection methods on various benchmarks and real-world datasets.
\end{itemize}

\section{Related Work}
We review representative noise-robust methods and sample selection strategies in learning with noisy labels (LNL), excluding studies that assume access to clean label subsets~\cite{clothing1m, hendrycks2018using, DAT, DMLP}.

\paragraph{Noise-robust methods.} These methods address noisy labels through robust loss functions, loss correction, label correction, and regularization.
\emph{Robust loss functions} like MAE~\cite{MAE}, GCE~\cite{GCE}, SCE\cite{SCE}, $\mathcal{L}_{\mathrm{DMI}}$\cite{L_DMI}, NCE\cite{NCE}, GJS\cite{GJS}, $f$-divergence\cite{f-divergence} and Peer loss\cite{peerloss} are designed to mitigate noise. 
\emph{Loss correction} methods estimate noise transition matrices, achieving success in class-dependent matrices estimating~\cite{CDTM, DualT}, though accurately estimating the more general instance-dependent noise transition matrix remains challenging without additional assumptions~\cite{PDTM, CSIDN, SCIDN, IFIDN, MRIDN, BLTM, TAIDTM}.
\emph{Label correction} replaces noisy labels with model outputs~\cite{JointOpt, PENCIL}. Bootstraping~\cite{Bootstraping} and M-correction~\cite{M-correction} use a convex combination of noisy labels and model predictions for training.
\emph{Regularization} methods constrain model capacity to prevent memorization. For instance, ELR~\cite{ELR} uses temporal regularization, NCR~\cite{NCR} enforces similarity among neighbors, and contrastive learning~\cite{C2D, Sel-Cl, CTTR, CRLNL, peng2023empirical} enhances robust representation. CS-Isolate~\cite{CS-isolate} disentangling style and content in the representation space, distancing hard samples from the decision boundary to ease learning. Other regularization techniques including MixUp~\cite{mixup}, label smoothing~\cite{LS, NLS, NERDenoise, Dialogue}, and early stopping~\cite{PES} further improve noise tolerance.

\paragraph{Sample selection methods.} These methods identify mislabeled samples using model predictions, representations, or training dynamics. 
\emph{Model prediction-based} methods typically regard samples with small losses as correctly labeled. Co-teaching~\cite{Co-teaching} and its variants~\cite{Co-teaching+, jocor, CNLCU} simultaneously train two collaborating networks, selecting small-loss samples for each other to reduce confirmation bias. These methods need to dynamically adjust the select ratio in each iteration, which can be tricky in practice. A more flexible method is fitting a two-component Beta/Gaussian Mixture Model (BMM/GMM) on per-sample loss to differentiate correct and incorrect labels~\cite{M-correction, DividMix, AugDesc, CC}.
\emph{Representation-based} approaches exploit latent features to distinguish clean from noisy data. CURST~\cite{CURST} selects samples that provide an approximately low-rank Jacobian matrix, which helps the network learn fast and generalize well. TopoFilter~\cite{TopoFilter} assumes clean data clusters together while corrupted data is spread out in the feature representation, using high-order topological information to identify correct labels. FINE~\cite{FINE} detects mislabeled samples through the principal components of latent representations made by eigendecomposition.
\emph{Training dynamics-based} methods, utilize model predictions over multiple iterations to generate more accurate sample selections. AUM~\cite{AUM} and recently proposed HMW~\cite{HMW} rank samples using the average logit margin to select correct labels. L2D~\cite{L2D} trains a noise detector based on training dynamics in a supervised manner, avoiding manual designing of the sample selection criterion. However, the pre-trained noise detector may not perform consistently well across different datasets with varying noise ratios, and fine-tuning the noise detector on target datasets requires additional clean data. DIST \cite{DISC} considered the fitting difficulty of different samples and proposed an instance-dependent sample selection criterion. Unlike previous methods that set a global or class-dependent threshold to select correct labels, they use the momentum maximum confidence of each instance computed across all previous epochs as the threshold value.

Our method belongs to sample selection methods based on training dynamics. Existing approaches typically set thresholds on per-sample loss, confidence, or logit margins to select correct labels, focusing primarily on the value in model predictions. These methods struggle to distinguish between correct and incorrect labels in high-loss data. In contrast, our method emphasizes the trend of confidence gaps in model predictions, enabling the identification of correct labels even within high-loss data. Combining our approach with existing methods can mitigate the trade-off between precision and recall in sample selection, selecting more correctly labeled yet hard-to-learn samples, resulting in stronger performance.

\section{Method}

\begin{figure*}
    \centering
    \includegraphics[width=0.95\textwidth]{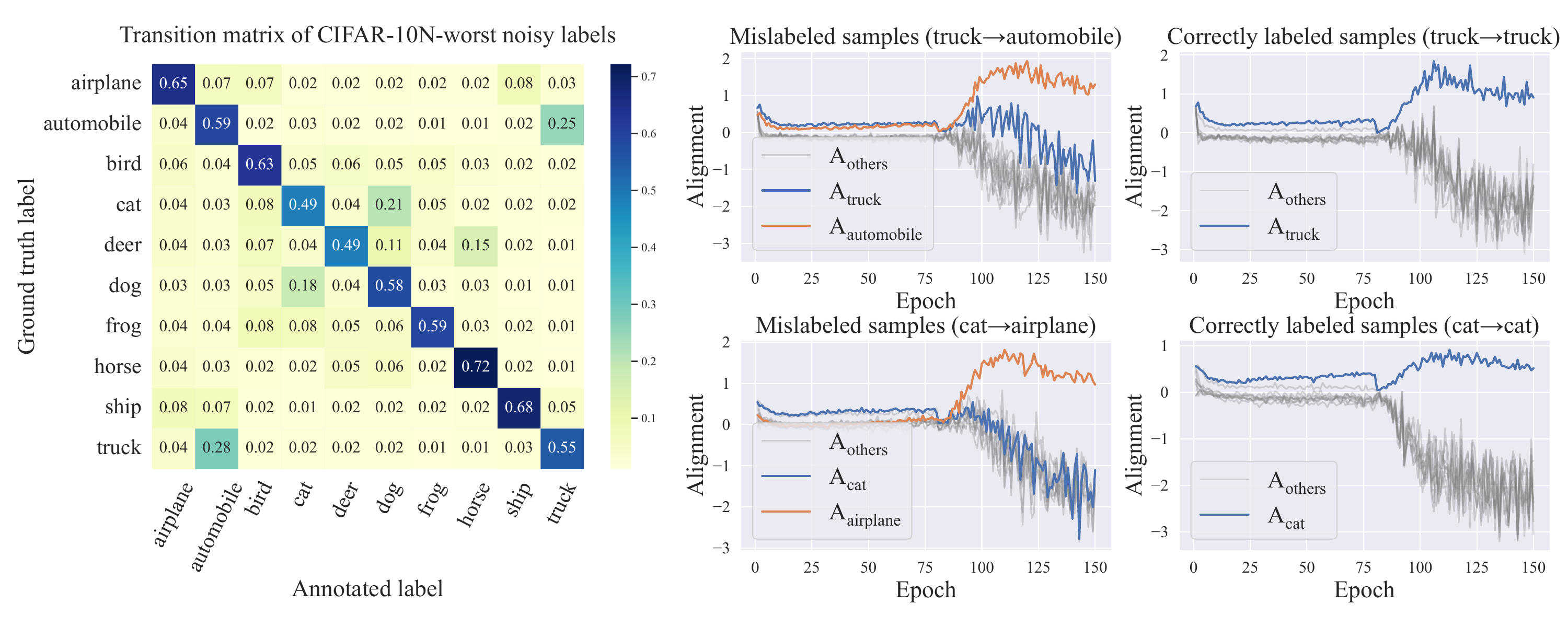}
    \caption{
    Left: Transition matrix of CIFAR-10N-Worst noisy labels. Right: Gradient alignment ( Equation~\ref{equation:alignment}) for different types of samples when training a PreActResNet-18 using cross-entropy loss and an SGD optimizer on CIFAR-10N-Worst.
    }
    \label{fig:gradient_alignment_CE}
\end{figure*}

\subsection{Problem Setup}
This paper considers the $k$-class classification problem in the presence of noisy labels. The noisy training set consists of $n$ examples $\mathcal{D}=\{\vec{x}^{[i]}, \vec{\hat{y}}^{[i]}\}_{i=1}^n$, where $\vec{x}^{[i]}\in \mathbb{R}^d$ is the $i$th input and $\vec{\hat{y}}^{[i]} \in \{0, 1\}^k$ is a one-hot vector indicating the annotated class. We use the non-bold letters $\hat{y}^{[i]}$ and $y^{[i]}$ to represent the annotated and ground truth classes of the $i$th input, respectively.
For simplicity, we denote a deep neural network as $f(\cdot;\vec{\theta}):\mathbb{R}^b \mapsto \mathbb{R}^k$, which maps $\vec{x}^{[i]}$ to the conditional probability $\vec{p}^{[i]}\in \mathbb{R}^k$ for each class. We use $f(\vec{x}^{[i]};\vec{\theta})_c$ to represent the conditional probability $\mathcal{P}(y^{[i]}=c \mid \vec{x}^{[i]})$ predicted by the model. Typically, the parameters $\vec{\theta} \in \mathbb{R}^p$ are optimized by minimizing the cross-entropy loss:
\begin{equation}
    \mathcal{L}_{\mathrm{CE}} = \frac{1}{n} \sum_{i=1}^{n} \ell_{\mathrm{CE}}(\hat{\vec{y}}^{[i]}, \vec{p}^{[i]}).
\end{equation}
\begin{equation}
    \ell_{\mathrm{CE}}(\hat{\vec{y}}^{[i]}, \vec{p}^{[i]}) = -\sum_{c=1}^{k} \hat{\vec{y}}^{[i]}_c \log{ f(\vec{x}^{[i]};\vec{\theta})_c}.
\end{equation}
We consider the common stochastic gradient descent method, where the dataset is divided into multiple mini-batches $\mathcal{B}=\{\mathrm{b}_{i}\}_{i=1}^{|\mathcal{B}|}$, and perform gradient descent on each batch:
\begin{equation}
    \mathcal{L}_{\mathrm{b}_t}(\vec{\theta}) = -\frac{1}{|\mathrm{b}_t|}\sum_{(\vec{x}^{[i]}, \vec{y}^{[i]})\in \mathrm{b}_t}\sum_{c=1}^{k} \hat{\vec{y}}^{[i]}_c \log{ f(\vec{x}^{[i]};\vec{\theta})_c},
\end{equation}
\begin{equation}\label{equation:gradientdescent}
    \vec{\theta}_{t+1} = \vec{\theta}_{t} - \eta g_t = \vec{\theta}_t - \eta \nabla \mathcal{L}_{\mathrm{b}_t}(\vec{\theta}_t),
\end{equation}
where $\eta$ denotes the learning rate and $\vec{\theta}_t, \mathrm{b}_t$ represent the parameters and the sampled mini-batch at timestep $t$, respectively. Following previous work~\cite{CORES, cleanlabeldominate}, we only consider \emph{clean-labels-dominant} dataset which means training samples are more likely to be annotated with true semantic labels than with any other class labels.

\subsection{An Empirical Analysis of Model Learning Process}\label{section:empirical_analysis}
Before detailing our method, we first analyze why the trend in confidence gaps can serve as a criterion for identifying correct labels in clean-labels-dominant datasets. We begin by examining how gradient descent over batches influences the classification model's prediction on a specific input. Typically, the parameters $\vec{\theta}$ do not change significantly in a single gradient descent step. Thus, the nonlinear function $f(\vec{x};\vec{\theta}_{t+1})_c$ can be approximated by its first-order Taylor expansion:
\begin{equation}\label{equation:taylor}
    f(\vec{x};\vec{\theta}_{t+1})_c \approx f(\vec{x}; \vec{\theta}_{t})_c + \langle \nabla_{\vec{\theta}_t}f(\vec{x};\vec{\theta}_t)_c, \vec{\theta}_{t+1} - \vec{\theta}_{t} \rangle.
\end{equation}
For input $\vec{x}$, changes in model's prediction confidence for class $c$ after a gradient descent step on mini-batch $\mathrm{b}_t$ can be modeled as follows:
\begin{equation}\label{equation:core}
    f(\vec{x};\vec{\theta}_{t+1})_{c} - f(\vec{x};\vec{\theta}_{t})_{c} \propto \langle \nabla_{\vec{\theta}_t}\ell_{\mathrm{CE}}(\vec{c}, \vec{x}) , \nabla \mathcal{L}_{\mathrm{b}_t}(\vec{\theta}_t) \rangle.
\end{equation}
The degree of alignment between $\nabla_{\vec{\theta}_t}\ell_{\mathrm{CE}}(\vec{c}, \vec{x})$ and $\nabla \mathcal{L}_{\mathrm{b}_t}(\vec{\theta}_t)$ determines the direction of change in the model's prediction confidence. Intuitively, if $\nabla \mathcal{L}_{\mathrm{b}_t}(\vec{\theta}_t)$ is aligned with $\nabla_{\vec{\theta}_t}\ell_{\mathrm{CE}}(\vec{c}, \vec{x})$, the gradient descent step on mini-batch $\mathrm{b}_t$ will decrease $\ell_{\mathrm{CE}}(\vec{c}, \vec{x})$ and increase the model's prediction confidence for class $c$ given input $\vec{x}$.

Previous studies on Coherent Gradients~\cite{coherentgradients} indicate that gradients from similar examples are alike, and the overall gradient is stronger in directions where these reinforce each other. Consider a randomly initialized model that outputs random guesses on all inputs. During the early stages of training, the model has not yet fit the given annotations, and the gradients from correct and incorrect labels typically have similar magnitudes. Additionally, the gradients from correct labels are coherent since correctly labeled samples usually share similar patterns. As a result, in the clean-labels-dominant datasets, the correct labels will dominate the overall gradients in the early training stage. This means $\nabla \mathcal{L}_{\mathrm{b}_t}(\vec{\theta}_t)$ tends to be most aligned with $\nabla_{\vec{\theta}_t}\ell_{\mathrm{CE}}(\vec{y}, \vec{x})$. In other words, the model's prediction confidence for ground truth labels tends to increase faster than for other classes in the early training stage. However, as the model gradually fits the correctly labeled examples, their gradients tend to diminish. Then the gradients from the under-fitted mislabeled examples will take over the gradient descent process, leading to the memorization of incorrect labels. To verify our analysis, we randomly sample a subset $\mathcal{N}=\{(\vec{x}^{[j]}, \vec{y}^{[j]}, \hat{\vec{y}}^{[j]})\}_{j=1}^{|\mathcal{N}|}$ consisting of examples from CIFAR-10N-Worst~\cite{CIFARN}, CIFAR-10 with noisy human annotations from Amazon Mechanical Turk, and report the following metrics in different training iterations:
\begin{equation}\label{equation:alignment}
    A_{c} = \frac{1}{|\mathcal{N}|} \frac{1}{|\mathcal{B}|} \sum_{t=1}^{|\mathcal{B}|} \sum_{j=1}^{|\mathcal{N}|} \langle \nabla_{\vec{\theta}_t}\ell_{\mathrm{CE}}(\vec{c}, \vec{x}^{[j]}) , \nabla \mathcal{L}_{\mathrm{b}_t}(\vec{\theta}_t) \rangle.
\end{equation}
Here, $t$ indicates the gradient descent step in one iteration, increasing from 1 to the number of batches in the dataset. $A_{c}$ measures the degree of alignment between $\nabla_{\vec{\theta}_t}\ell_{\mathrm{CE}}(\vec{c}, \vec{x})$ and the gradients over batches in one iteration. The larger $A_c$ is, the faster the model's prediction confidence increases in category $c$. Figure~\ref{fig:gradient_alignment_CE} shows the transition matrix of CIFAR-10N-worst noisy labels and the trajectories of $A_c$ on different kinds of samples. The experimental results are consistent with our analysis: $A_{y}$ is highest on both correctly labeled and mislabeled samples in the early training stage. Only after a period of training, $A_{\hat{y}}$ will have a larger value than $A_{y}$. These empirical findings align with the \emph{early learning} phenomenon~\cite{ELR}, also known as memorization effect~\cite{memorizationeffect}, which suggests the deep neural networks optimized by SGD typically learn from correct labels before overfitting noisy data. This phenomenon has already been proved under high-dimensional linear classification~\cite{ELR}. We provide additional evidence to illustrate why this occurs in deep neural networks. Note this phenomenon only suggests that the model's prediction confidence for ground truth labels usually rises fastest among all classes in the early training stage. However, it does not guarantee that the model will output high-confidence predictions for ground truth labels since some samples may be difficult to fit, causing the confidence to rise slowly and resulting in correctly labeled examples with high loss, just like the hard-to-learn dogs' images in Figure~\ref{fig:trainingdynamics}. This motivates us to design a novel sample selection method that considers trends in the model's prediction confidence rather than relying solely on loss values.

\subsection{Sample Selection by Confidence Tracking}\label{section:ct_details}
Our observations indicate that during the early stages of training, only for the correctly labeled samples, the model's prediction confidence for annotated labels usually increases faster than for any other classes. Based on this phenomenon, we introduce a novel sample selection method. Generally, for an input $\vec{x}^{[i]}$, if the model's prediction confidence increases faster for class $c_1$ than for class $c_2$, then the confidence gap between $c_1$ and $c_2$ should increase:
\begin{align}
    \begin{split}
        &\vec{p}^{[i]}_{c_1}(t+1) - \vec{p}^{[i]}_{c_1}(t) > \vec{p}^{[i]}_{c_2}(t+1) - \vec{p}^{[i]}_{c_2}(t) \\
        \Rightarrow & \vec{p}^{[i]}_{c_1}(t+1) - \vec{p}^{[i]}_{c_2}(t+1) > \vec{p}^{[i]}_{c_1}(t) - \vec{p}^{[i]}_{c_2}(t),
    \end{split}
\end{align}
where $\vec{p}^{[i]}_{c}(t)$ denotes the model's prediction confidence for $\vec{x}^{[i]}$ in class $c$ at iteration $t$. If the confidence gaps between the annotated label and other labels all tend to increase, the confidence should rise fastest on the annotated label. We regard such samples as potentially correctly labeled. To implement this idea, for each example $(\vec{x}^{[i]}, \hat{\vec{y}}^{[i]})$ in the noisy training set, we first collect the following confidence gaps:
\begin{equation}
    d^{[i]}_c(t) = \vec{p}^{[i]}_{\hat{y}^{[i]}}(t) - \vec{p}^{[i]}_{c}(t).
\end{equation}
Gathering those confidence gaps over different training iterations, we obtain the following series:
\begin{equation}
    D^{[i]}_c(t) = \{d^{[i]}_c(1), d^{[i]}_c(2), ...,  d^{[i]}_c(t)\}.
\end{equation}
To judge the trend of these series, we utilize the Mann-Kendall Trend Test~\cite{mann1945, kendall1975}.
This is a non-parametric method based on the ranks of the data rather than their actual values, making it robust against extreme values. We use $\text{MK-Test}(\cdot)$ to represent the Mann-Kendall testing process, which takes a series of data as input and outputs the standardized test statistic $Z$. Our alternative hypothesis points to an upward trend in the confidence gaps series. Formally, our sample selection criterion is:
\begin{equation}\label{equation:criterion}
    \underbrace{\underset{c\in \{1, 2, ..., k\}\backslash \{\hat{y}^{[i]}\}}{\text{min}} \text{MK-Test}(D^{[i]}_c(t))}_{Z_{min}^{[i]}} > Z_{1-\alpha},
\end{equation}
where $\alpha$ is the chosen significance level and $Z_{1 - \alpha}$ is the $100(1-\alpha)$th percentile of the standard normal distribution. If $\text{MK-Test}(D^{[i]}_c(t)) > Z_{1-\alpha}$, the probability of $D^{[i]}_c(t)$ has no trend is less than $\alpha$ so we accept the alternative hypothesis. Therefore, if an example $(\vec{x}^{[i]}, \vec{\hat{y}}^{i})$ satisfies Equation~\ref{equation:criterion} at iteration $t$, it suggests that all confidence gaps have an upward trend, so we regard it as potentially correctly labeled.

In practice, we combine Confidence Tracking~(CT) with existing sample selection methods to enhance their performance. Specifically, we use the union of samples selected by CT and other methods for training. Current methods reliably select correct labels from small-loss samples, while CT can identify correct labels from high-loss samples. This combination aims to maintain precision and improve recall for sample selection. Let $C_t$ denote the selected examples at iteration $t$, we assign zero weight to samples not in subset $C_t$ in the loss function:
\begin{equation}\label{equation:loss2}
    \mathcal{L}_{t} = \frac{1}{n} \sum_{i=1}^{n} \mathbb{I}((\vec{x}^{[i]}, \hat{\vec{y}}^{[i]}) \in \mathcal{C}_{t}) \ell_{\mathrm{CE}}(\hat{\vec{y}}^{[i]}, \vec{p}^{[i]}).
\end{equation}
Following the common practice in sample selection~\cite{DividMix, FINE, DISC}, we first warm up the network using standard cross-entropy loss for several epochs. Then we begin to select a potentially clean subset at the end of each epoch and apply Equation~\ref{equation:loss2} for training. Generally, the selected subset likely excludes some mislabeled data, aligning the overall gradient towards memorizing ground truth labels, thereby promoting correct label memorization. This helps to generate better sample selection results in the following iterations. Such a positive feedback loop gradually eliminates mislabeled data from the noisy training set. Section~\ref{appendix:alg} in the technical appendix provides the details of the Mann-Kendall Trend Test and pseudo-code of our algorithm.

\section{Experiment}

\begin{table*}[htbp]
  \centering
  \setlength{\tabcolsep}{1mm}
    \begin{tabular}{lccccccccc}
    \toprule
    Dataset & \multicolumn{4}{c}{CIFAR-10}  & \multicolumn{4}{c}{CIFAR-100} & \multirow{2}[4]{*}{Avg} \\
\cmidrule{1-9}    Noise type & Sym. 20\% & Sym. 50\% & Asym. 40\% & Real. 40\% & Sym. 20\% & Sym. 50\% & Asym. 40\% & Real. 40\% &  \\
    \midrule
    CE    & 86.51±0.22 & 77.41±0.65 & 83.78±1.76 & 77.94±0.91 & 61.37±0.12 & 46.82±1.32 & 45.70±0.39 & 52.82±0.30 & 66.54  \\
    L2D   & 92.25±0.12 & 87.27±0.55 & 82.57±1.31 & 84.50±0.44 & 71.05±0.47 & 60.82±0.59 & 47.94±0.63 & 59.44±0.33 & 73.23  \\
    Co-teaching & 91.88±0.21 & 87.58±0.41 & 87.72±1.00 & 85.22±0.28 & 70.45±0.36 & 64.07±0.47 & 58.95±0.91 & 62.32±0.26 & 76.03  \\
    CNLCU & 91.92±0.36 & 87.58±0.73 & 88.14±0.61 & 86.08±0.39 & 70.61±0.34 & 63.87±0.11 & 55.94±0.73 & 62.28±0.20 & 75.80  \\
    HMW   & 92.02±0.21 & 87.81±0.22 & 87.37±0.29 & 85.28±0.29 & 72.01±0.22 & 65.22±0.36 & 64.69±0.13 & 61.52±0.28 & 76.99  \\
    \midrule
    GMM   & 91.60±0.28 & 88.07±0.08 & 89.45±0.85 & 86.63±0.34 & 69.59±0.32 & 63.95±0.44 & 65.29±0.42 & 60.22±0.23 & 76.85  \\
    GMM+CT & \textbf{92.57±0.12} & \textbf{89.11±0.21} & \textbf{90.55±0.22} & \textbf{87.33±0.38} & \textbf{71.37±0.67} & \textbf{65.17±0.39} & \textbf{68.84±0.47} & \textbf{62.73±0.14} & 78.46  \\
    \midrule
    FINE  & 89.13±0.48 & 85.66±0.32 & 82.56±2.00 & 80.09±0.45 & 70.96±0.45 & 58.58±0.48 & 49.48±0.77 & 56.87±0.25 & 71.67  \\
    FINE+CT & \textbf{92.48±0.21} & \textbf{87.56±0.13} & \textbf{86.92±0.51} & \textbf{84.22±0.49} & \textbf{71.17±0.32} & \textbf{58.76±0.41} & \textbf{53.16±0.88} & \textbf{58.52±0.15} & 74.10  \\
    \midrule
    AUM   & 92.31±0.13 & 87.80±0.24 & 88.21±0.54 & 86.22±0.11 & 72.50±0.44 & 64.90±0.28 & 61.25±0.41 & 61.75±0.38 & 76.87  \\
    AUM+CT & \textbf{92.45±0.13} & 87.91±0.40 & \textbf{89.70±0.40} & \textbf{87.29±0.16} & 72.56±0.18 & 64.99±0.41 & \textbf{63.80±0.37} & \textbf{62.05±0.18} & 77.59  \\
    \midrule
    DIST  & 92.63±0.15 & 88.43±0.24 & 90.00±0.42 & 86.39±0.54 & 72.73±0.32 & 65.59±0.24 & 66.74±0.81 & 60.97±0.20 & 77.93  \\
    DIST+CT & 92.43±0.31 & 88.35±0.16 & \textbf{90.58±0.21} & \textbf{87.01±0.43} & 72.78±0.27 & 65.51±0.16 & \textbf{69.05±0.47} & \textbf{62.18±0.17} & 78.49  \\
    \bottomrule
    \end{tabular}
  \caption{Test accuracy (\%) of different methods on CIFAR-10 and CIFAR-100 with symmetric, asymmetric, and real-world noisy labels (CIFAR-10N-Worst and CIFAR-100N-Noisy). We implement all methods based on public code and report mean accuracy and standard deviation over five random seeds. We use the Wilcoxon signed-rank test with a confidence level of 0.05 to compare the performance and bold the results where CT brings significant improvements.}
    \label{tab:main_results}
\end{table*}

\begin{table*}[ht!]
  \centering
  \setlength{\tabcolsep}{1.2mm}
    \begin{tabular}{lccccccccccccccc}
    \toprule
    Noise setting & \multicolumn{3}{c}{Sym. 20\%} & \multicolumn{3}{c}{Sym. 50\%} & \multicolumn{3}{c}{Asym. 40\%} & \multicolumn{3}{c}{Real. 40\%} & \multicolumn{3}{c}{Average} \\
    \midrule
    Metric & P     & R     & F1    & P     & R     & F1    & P     & R     & F1    & P     & R     & F1    & P     & R     & F1 \\
    \midrule
    GMM   & 99.86  & 71.11  & 83.06  & 98.97  & 69.46  & 81.62  & 98.64  & 63.84  & 77.51  & 89.52  & 72.82  & 80.31  & 96.75  & 69.30  & 80.62  \\
    GMM+CT & 99.71  & 83.36  & 90.80  & 98.33  & 79.61  & 87.99  & 96.46  & 82.12  & 88.72  & 87.04  & 84.61  & 85.81  & 95.39  & 82.43  & \textbf{88.33} \\
    \midrule
    FINE  & 97.93  & 93.23  & 95.52  & 85.68  & 95.14  & 90.16  & 69.98  & 72.13  & 71.03  & 73.23  & 83.68  & 78.11  & 81.70  & 86.05  & 83.70  \\
    FINE+CT & 97.65  & 96.34  & 96.99  & 85.52  & 96.16  & 90.53  & 72.76  & 85.90  & 78.78  & 72.97  & 93.99  & 82.15  & 82.22  & 93.10  & \textbf{87.11} \\
    \midrule
    AUM   & 99.09  & 92.84  & 95.86  & 96.15  & 86.67  & 91.17  & 86.26  & 72.03  & 78.51  & 84.48  & 84.67  & 84.57  & 91.50  & 84.05  & 87.53  \\
    AUM+CT & 99.10  & 92.99  & 95.95  & 96.14  & 87.06  & 91.38  & 88.82  & 76.69  & 82.31  & 84.58  & 85.71  & 85.14  & 92.16  & 85.61  & \textbf{88.69} \\
    \midrule
    DIST  & 99.04  & 94.74  & 96.84  & 95.56  & 91.56  & 93.51  & 96.87  & 73.88  & 83.81  & 85.80  & 79.10  & 82.31  & 94.32  & 84.82  & 89.12  \\
    DIST+CT & 99.00  & 94.93  & 96.93  & 95.27  & 92.25  & 93.73  & 96.56  & 87.55  & 91.84  & 85.88  & 86.32  & 86.09  & 94.18  & 90.27  & \textbf{92.15} \\
    \bottomrule
    \end{tabular}%
  \caption{Comparisons of sample selection precision, recall, and F1-score to correct labels on CIFAR-100 dataset with symmetric, asymmetric, and real-world noisy labels. Results are averaged over five random seeds.}
  \label{tab:c100_p_r_f1}
\end{table*}

\subsection{Experimental settings}
We evaluate our approach on four benchmarks, namely CIFAR-10, CIFAR-100~\cite{CIFAR}, WebVision~\cite{webvision}, and Food-101N~\cite{food101n}. For CIFAR-10 and CIFAR-100, we experiment with both simulated and human-annotated real-world noisy labels. For simulated noise, we follow the previous setups~\cite{CDTM, ELR} and experiment with two types of label noise: \emph{symmetric} and \emph{asymmetric}. Symmetric noise is generated by randomly replacing the original labels with all other possible classes. Asymmetric noise is a more realistic setting where labels are replaced by similar classes. We generate asymmetric noise following the same schema with previous works~\cite{ELR, FINE}. For CIFAR-10, we map Turck $\to$ Automobile, Bird $\to$ Airplane, Deer $\to$ Horse, Cat $\leftrightarrow$ Dog. For CIFAR-100, we create 20 five-size super-classes and replace the original label with the next class within super-classes circularly. For human-annotated real-world noise, we experiment with the CIFARN~\cite{CIFARN} dataset (CIFAR-10/100 dataset with noisy human annotations from Amazon Mechanical Turk). Unlike simulated class-dependent noise, real-world noise patterns are instance-dependent making it more challenging to identify correct labels. 

WebVision and Food-101N are two real-world noisy datasets. WebVision dataset contains 2.4 million images crawled from the web and its estimated noise rate is about 20\%~\cite{webvision}. Following previous work~\cite{ELR}, we use the mini WebVision dataset for training, which contains the first 50 classes from the Google image subset (about 66 thousand images), and evaluate model performance on both WebVision and ImageNet ILSVRC12 validation sets~\cite{imagenet_cvpr09} using InceptionResNetV2~\cite{InceptionResNet}. The Food-101N dataset contains about 310,009 images of food recipes classified in 101 categories and its estimated noise rate is about 20\%. Food-101N and the Food-101 dataset~\cite{food101} share the same 101 classes, whereas Food-101N has much more images and is more noisy. Following previous work~\cite{food101n}, we use ResNet50~\cite{ResNet} pretrained on ImageNet~\cite{imagenet_cvpr09} and evaluate model performance on the Food-101 test set.

\subsection{Integrating with Sample Selection Methods}\label{sec:main_results}
We mainly experiment with the following advanced sample selection methods: Co-teaching~\cite{Co-teaching}, CNLCU~\cite{CNLCU}, GMM~\cite{DividMix}, FINE~\cite{FINE}, L2D~\cite{L2D}, AUM~\cite{AUM}, DIST~\cite{DISC}, and HMW~\cite{HMW}, which cover representative methods based on model predictions, sample representations, and training dynamics. Section~\ref{appendix:train_details} in the technical appendix provides a more detailed introduction and implementation details of those baselines.

\begin{figure*}
    \centering
    \includegraphics[width=\textwidth]{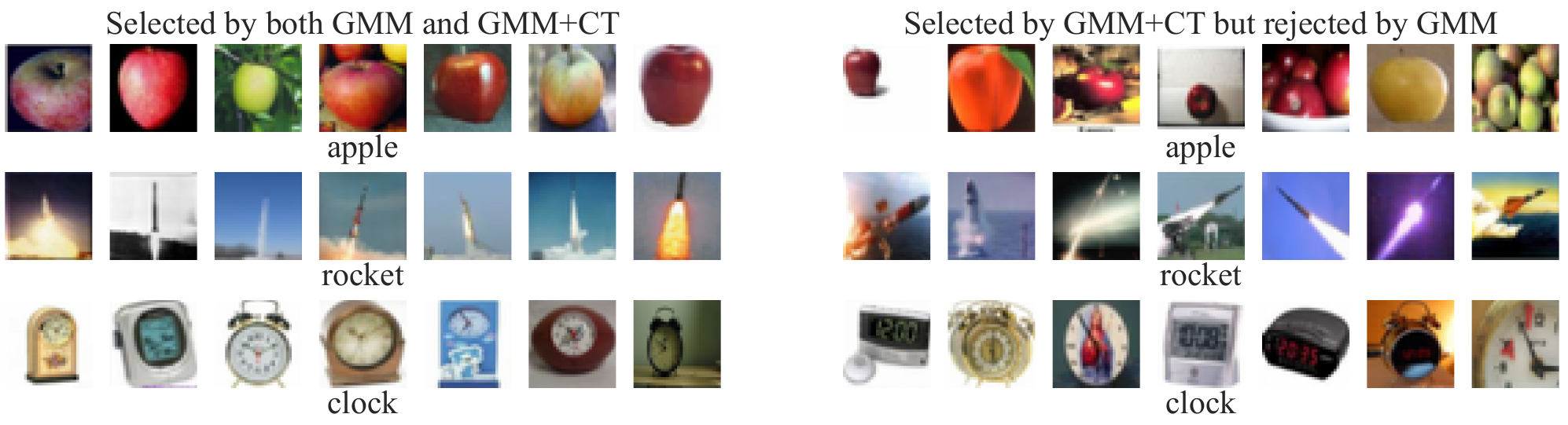}
    \caption{The sample selection results on the CIFAR-100N-noisy dataset. The left graph shows samples selected by both GMM and GMM+CT and the right graph shows samples selected by GMM+CT but rejected by GMM. These samples are chosen randomly, not cherry-picked.}
    \label{fig:case_study}
\end{figure*}

Tabel~\ref{tab:main_results} demonstrates the performance when integrating CT with state-of-the-art sample selection approaches. All methods use the same architecture (PreActResNet-18) and training procedure (please refer to Section~\ref{appendix:train_details} in the technique appendix for more details). The significance level $\alpha$ used in CT is set to 0.01 and the warm-up epoch is 30 for all methods. We retain 10\% of the training sets to perform validation and select the model with the best validation performance for the test. CT brings consistent improvement to various baselines. We notice the improvement is more significant under asymmetric and real-world noise. This is because the symmetric noise randomly changes the correct label to another one, making the gradient between noise samples usually incoherent. As a result, the model fits correct labels much faster than incorrect ones, making it easy to distinguish them using the loss values. So sample selection methods based on the small-loss criterion can achieve satisfying performance under symmetric noise. However, under asymmetric or real-world noise, noise labels are more structured (\eg trucks are generally mislabeled as automobiles rather than other categories), meaning the gradient of this mislabeled data is also coherent. Thus, the model can quickly fit mislabeled data, making the loss of some correctly labeled yet hard-to-learn samples similar to mislabeled ones. The popular small-loss criterion is less effective under such situations, while CT can still distinguish hard correctly labeled data from mislabeled ones. Tabel~\ref{tab:c100_p_r_f1} compares the sample selection precision and recall for correct labels of different methods. Generally, integrating CT with other sample selection methods significantly improves recall and maintains the precision of sample selection which shows CT accurately identifies correct labels from samples rejected by existing sample selection methods, \ie samples with relatively high loss or logit margins. These results confirm that CT is more powerful than existing sample selection methods in distinguishing correctly labeled yet hard-to-learn and mislabeled samples. Section~\ref{appendix:additional_results} in the technical appendix provides additional results on synthetic instance-dependent label noise~\cite{PDTM}.

Figure~\ref{fig:case_study} compares the sample selected by GMM and GMM+CT on the CIFAR-100N-noisy dataset. Compared with images selected by both GMM and GMM+CT, images selected only by GMM+CT show greater inter-class variability. For instance, the apple images encompass different environment settings and perspectives; the rocket images capture various stages of rocket launches; the clock images display diverse designs and time formats. It intuitively shows that introducing CT can select a richer sample set, which helps improve model performance. Due to the page limit, we further analyze the robustness of CT in Section~\ref{appendix:robustness} in the technical appendix, which shows CT is not sensitive to the choice of $\alpha$ and the number of warm-up epochs.

\begin{table}[t!]
  \centering
  \setlength{\tabcolsep}{1mm}
    \begin{tabular}{lcccccc}
    \toprule
    Test dataset & \multicolumn{2}{c}{Webvision} & \multicolumn{2}{c}{ILSVRC12} & Food-101N \\
    \midrule
    Metric & Top1  & Top5  & Top1  & Top5  & ACC   \\
    \midrule
    CORES & 71.70  & 89.02  & 68.36  & 88.28  & 84.38   \\
    CORES+CT & \textbf{72.11} & \textbf{90.24} & \textbf{68.52} & \textbf{90.03} & \textbf{84.43} \\
    \midrule
    DivideMix & 77.44  & 91.88  & 74.72  & 92.12  & 86.53  \\
    DivideMix+CT & \textbf{78.12} & \textbf{92.26} & \textbf{75.23} & \textbf{92.19} & \textbf{86.76} \\
    \midrule
    f-DivideMix & 78.36  & 92.54  & 75.25  & 92.22  & 86.83  \\
    f-DivideMix+CT & \textbf{78.81} & \textbf{92.91} & \textbf{75.66} & \textbf{93.22} & \textbf{87.05} \\
    \midrule
    DISC  & 80.07  & 92.38  & 77.40  & 92.38  & 87.32  \\
    DISC+CT & 80.07  & \textbf{92.56} & \textbf{78.26} & \textbf{92.43} & \textbf{87.45} \\
    \bottomrule
    \end{tabular}%
    \caption{The average test accuracy (\%) over the last 10 epochs on the real-world noisy dataset.}
  \label{tab:SOTA+CT}%
\end{table}%

\subsection{Integrating with Advanced LNL Methods}
Existing state-of-the-art methods of learning with noisy labels usually combine sample selection with semi-supervised learning to further improve performance. In this section, we integrate CT with CORES~\cite{CORES}, DivideMix~\cite{DividMix}, f-DivideMix~\cite{FINE}, and DISC~\cite{DISC} to analyze whether our sample selection procedure can bring further improvement to these state-of-the-art methods. Table~\ref{tab:SOTA+CT} shows CT consistently improves the performance of all baselines on real-world noisy datasets.

\section{Conclusion}
In this paper, we introduced a novel sample selection method for image classification with noisy labels, termed Confidence Tracking (CT). Unlike existing methods that rely on small-loss criteria, our approach leverages the observation that only for the correctly labeled samples, the model's prediction confidence for annotated labels usually increases faster than for any other classes. By monitoring the trends in confidence gaps between annotated labels and other classes, CT effectively distinguishes correctly labeled samples even when they exhibit high losses during training. Our experimental results demonstrate that CT enhances the performance of existing learning-with-noisy-labels (LNL) methods across various benchmarks, showcasing its robustness and reliability. This method successfully alleviates the trade-off dilemma between precision and recall in sample selection when setting a threshold on pre-sample loss, confidence, or logit margins, offering a more accurate identification of hard-to-learn yet correctly labeled samples. Future research could explore a deeper theoretical understanding of the early learning phenomenon and the development of more advanced sample selection criteria based on training dynamics. 

\section*{Acknowledgments}
This work was supported in part by the National Natural Science Foundation of China under Grant No. 62276110, No. 62172039, and in part by the fund of Joint Laboratory of HUST and Pingan Property \& Casualty Research (HPL). The authors would also like to thank the anonymous reviewers for their comments on improving the quality of this paper.

\bibliography{aaai25}

\clearpage
\appendix

\section{Overview of Technical Appendix}
Section~\ref{appendix:dataset} provides information on the benchmark datasets we used. Section~\ref{appendix:alg} provides implementation details and pseudocode for CT. Section~\ref{appendix:additional_results} shows additional experiments. Section~\ref{appendix:robustness} analyzes the robustness of CT. Section~\ref{appendix:train_details} gives detailed descriptions of our experiments. Section~\ref{appendix:further_discussion} explores whether sample selection methods can further benefit from strong CLIP features.
Section~\ref{appendix:limitations} discusses limitations and broader impacts of our method.

\section{Dataset Information}\label{appendix:dataset}
We use CIFAR-10~\cite{CIFAR}, CIFAR-100~\cite{CIFAR}, CIFARN~\cite{CIFARN}, WebVision~\cite{webvision}, and Food101-N~\cite{food101n} in our experiments. For CIFAR-10, CIFAR-100, and CIFARN, we retain 10\% of the training set for validation and select the model with the best validation performance for the test. For WebVision, we only consider examples belonging to the top 50 classes from the Google image subset, namely miniWebVision, and evaluate model performance on both WebVision and ImageNet ILSVRC12 validation sets~\cite{imagenet_cvpr09}. For Food101-N, we evaluate model performance on the test set of Food-101~\cite{food101}. We do not observe overfitting in our experiments on WebVision and Food101N and report the average test accuracy over the last 10 epochs. The licenses of CIFAR-10, CIFAR-100, Food-101, and Food-101N are not explicitly stated but they are generally available for non-commercial research purposes. CIFARN is licensed under CC BY-NC 4.0. WebVision and ImageNet ILSVRC12 have custom licenses and are available for non-commercial research purposes.

\begin{algorithm}[t!]
    \caption{Pseudocode for GMM+CT.}
    \label{alg:ct}
    \DontPrintSemicolon
    \textbf{Input:} training data with noisy labels $\mathcal{D}=\{\ns \}_{i=1}^{N}$; number of classes $K$; numbers of training epochs $T$; a neural network $f(\cdot; \vec{\theta})$ with trainable parameters $\vec{\theta}$; significance level $\alpha$ used in CT.
    
    $gaps\gets \vec{0}_{[N\times K \times T]}$ \\
    $S \gets \vec{0}_{[N]}$  \\
    $loss \gets \vec{0}_{[N]}$ \\
    $\mathcal{C} \gets \{i \mid i \in \mathbb{N}, 1 \leq i \leq N\}$ \\
    
    \For{$e$ in $[1,\mathit{num\_epochs}]$}{
        \ForEach{minibatch $B=\{idx, imgs, labels\}$}{
            $out \gets f(imgs; \theta)$ \\
            $Update\_S(out, idx, S, gaps, e)$ \\
            $\mathcal{L} \gets \frac{1}{|B|} \sum_{B} \mathbb{I}(idx \in \mathcal{C}) \ell_{\mathrm{CE}}(out, labels)$ \\
            $loss[idx]\gets\mathcal{L}$ \\
            
            $\vec{\theta} = SGD(\mathcal{L}, \vec{\theta}$) \\
        }
        \If{$e \geq warmup\_epochs$}{
            $\mathcal{C}_{GMM} \gets GMM(loss)$
            $\mathcal{C}_{CT} \gets CT(S, e, \alpha)$
            $\mathcal{C} \gets \mathcal{C}_{GMM} \cup \mathcal{C}_{CT}$
        }
    }   
\end{algorithm}

\section{Algorithms Details}\label{appendix:alg}
\paragraph{Details of Mann-Kendall trend test.} This is a non-parametric method to detect trends in a series of data. It is based on the ranks of the data rather than their actual values, making it robust against extreme values. Specifically, this method calculates the test statistic $S$ as follows:
\begin{equation}\label{equation:cal_s}
    S = \sum_{j=2}^{n} \sum_{k=1}^{j-1} \mathrm{sgn}(x_j - x_k),
\end{equation}
where $n$ is length of the series and $\mathrm{sgn}(x_j-x_k)$ is the sign function:
\begin{equation}
    \mathrm{sgn}(x_j - x_k) = 
    \begin{cases} 
    1 & \text{if } x_j > x_k \\
    0 & \text{if } x_j = x_k \\
    -1 & \text{if } x_j < x_k 
    \end{cases}
\end{equation}
$S$ analyzes the sign differences between later and earlier data points in the series. If the time series is monotonically increasing, $S$ will be positive, while a monotonically decreasing series gives a negative $S$.
The null hypothesis of the test is that there is no trend, which means that the observations are randomly ordered in the series. If $S$ is significantly different from zero, we reject the null hypothesis and conclude that there is a trend. The variance of $S$ under the null hypothesis is given by:
\begin{equation}
    \text{Var}(S) = \frac{1}{18} \left(n(n-1)(2n+5)\right),
\end{equation}
where $n$ is the number of data points. To facilitate implementation, we do not consider the tied ranks here. Finally, the standardized test statistic $Z$ is calculated as:
\begin{equation}\label{equation:get_z}
    Z = 
    \begin{cases} 
    \frac{S-1}{\sqrt{\text{Var}(S)}} & \text{if } S > 0 \\
    0 & \text{if } S = 0 \\
    \frac{S+1}{\sqrt{\text{Var}(S)}} & \text{if } S < 0 
    \end{cases}
\end{equation}

\paragraph{Pseudocode.} Algorithm~\ref{alg:ct} provides detailed pseudocode for GMM+CT. 
Specifically, Line 2 to 5 initializes confidence gaps, MK-Test statistic $S$, per-sample loss, and index of the selected subset $\mathcal{C}$, respectively. Lines 6 to 13 describe the training loop, where line 8 obtains the network output, line 9 updates the MK-Test statistic $S$, line 10 calculates training loss only on selected samples, line 11 saves per-sample loss, and line 12 updates the model parameters by SGD. Lines 14 to 18 describe the sample selection procedure, which uses the union of potentially clean subsets selected by GMM and CT to update $\mathcal{C}$. Note Line 14 shows we only perform sample selection after the warm-up period. During warm-up, all the training sample is used for training. Other variants like FINE+CT, AUM+CT, and DISC+CT have similar implementations. For example, replacing GMM in Algorithm~\ref{alg:ct} with DIST will give DIST+CT. The key to implementing CT is calculating the MK-Test statistic $S$. Then we can perform sample selection according to Equation~\ref{equation:get_z} and Equation~\ref{equation:criterion}. $S$ can be easily updated on the fly during training. Algorithm~\ref{alg:update_s} shows our implementation.

\begin{algorithm*}[htbp]
    \caption{Pseudocode to update statistic $S$ in PyTorch-like style.}
    \label{alg:update_s}
    \DontPrintSemicolon
    \# N: number of samples; K: number of classes; T: numbers of training epochs; B: batch size; \\
    \# e: current epoch; S: statistic $S$ in shape (N, K); gaps: confidence gaps in shape (N, T, K); \\
    \# out: model predictions in shape (B, K); index: sample index in shape (B,); \\
    \# label: annotated label in shape (B,); \\

    \text{  }\\
    
    \# get confidence gap between annotated label and others \\
    \texttt{g = out[torch.arange(B), label] - out} \\

    \text{  }\\

    \# save current confidence gap \\
    \texttt{gaps[index, e] = g}

    \text{  }\\

    \# update statistic $S$ \\
    \texttt{g = g.unsqueeze(1)} \\
    \texttt{delta = g.gt(gaps[index, :e]).sum(1) - g.lt(gaps[index, :e]).sum(1)} \\
    \texttt{S[index] = S[index] + delta} \\ 
\end{algorithm*}

\begin{table*}[t]
  \centering
  \setlength{\tabcolsep}{1mm}
    \begin{tabular}{lccccccccc}
    \toprule
    Dataset & \multicolumn{6}{c}{CIFAR-10N}                 & \multicolumn{2}{c}{CIFAR-100N} & \multirow{2}[4]{*}{Avg} \\
\cmidrule{1-9}    Noise type & Clean & Aggre & Rand1 & Rand2 & Rand3 & Worst & Clean & Noisy &  \\
    \midrule
    CE    & 94.45±0.06 & 89.76±0.09 & 86.38±0.24 & 85.94±0.55 & 85.86±0.17 & 77.94±0.91 & 74.93±0.22 & 52.82±0.30 & 81.01  \\
    \midrule
    GMM   & 92.85±0.11 & 91.77±0.18 & 91.03±0.12 & 90.82±0.14 & 90.96±0.16 & 86.63±0.34 & 68.91±0.27 & 60.22±0.23 & 84.15  \\
    GMM+CT & \textbf{93.49±0.16} & \textbf{92.68±0.09} & \textbf{91.83±0.22} & \textbf{91.83±0.26} & \textbf{92.02±0.27} & \textbf{87.33±0.38} & \textbf{73.56±0.45} & \textbf{62.73±0.14} & 85.68  \\
    \midrule
    FINE  & 89.29±0.24 & 88.99±0.45 & 88.26±0.49 & 88.41±0.31 & 88.39±0.47 & 80.09±0.45 & 71.46±0.53 & 56.87±0.25 & 81.47  \\
    FINE+CT & \textbf{93.11±0.12} & \textbf{92.51±0.22} & \textbf{91.59±0.19} & \textbf{91.86±0.18} & \textbf{91.86±0.16} & \textbf{84.22±0.49} & \textbf{74.13±0.20} & \textbf{58.52±0.15} & 84.72  \\
    \midrule
    AUM   & 93.73±0.09 & 92.23±0.18 & 91.07±0.29 & 91.33±0.22 & 91.12±0.16 & 86.22±0.11 & 75.29±0.28 & 61.75±0.38 & 85.34  \\
    AUM+CT & 93.86±0.14 & \textbf{92.54±0.10} & \textbf{91.77±0.24} & \textbf{91.81±0.13} & \textbf{91.64±0.19} & \textbf{87.29±0.16} & 75.26±0.18 & \textbf{62.05±0.18} & 85.78  \\
    \midrule
    DIST  & 94.21±0.27 & 92.43±0.31 & 91.43±0.16 & 91.65±0.23 & 91.50±0.24 & 86.39±0.54 & 75.33±0.14 & 60.97±0.20 & 85.49  \\
    DIST+CT & 94.31±0.07 & 92.58±0.29 & \textbf{91.79±0.18} & \textbf{91.96±0.22} & \textbf{91.90±0.25} & \textbf{87.01±0.43} & 75.15±0.34 & \textbf{62.18±0.17} & 85.86  \\
    \bottomrule
    \end{tabular}
  \caption{Comparison with SOTA sample selection methods in test accuracy (\%) on CIFARN dataset. The results are averaged over five random seeds. We use Wilcoxon signed-rank test with a confidence level of 0.05 to compare the performance and bold the results where CT brings significant improvements.}
  \label{tab:cifarn}
\end{table*}

\begin{table*}[t!]
  \centering
  \setlength{\tabcolsep}{1mm}
    \begin{tabular}{lccccccc}
    \toprule
    Dataset & \multicolumn{3}{c}{CIFAR-10} & \multicolumn{3}{c}{CIFAR-100} & \multirow{2}[4]{*}{Avg} \\
\cmidrule{1-7}    Noise type & IDN. 10\% & IDN. 30\% & IDN. 50\% & IDN. 10\% & IDN. 30\% & IDN. 50\% &  \\
    \midrule
    GMM   & 92.51±0.19 & 90.85±0.23 & 74.56±5.37 & 69.70±0.20 & 68.36±0.27 & 61.33±0.60 & 76.22  \\
    GMM+CT & \textbf{93.70±0.16} & \textbf{92.53±0.16} & \textbf{82.17±6.32} & \textbf{73.80±0.23} & \textbf{71.29±0.29} & \textbf{62.80±0.65} & 79.38  \\
    \midrule
    AUM   & 93.39±0.25 & 91.51±0.25 & 75.25±5.16 & 73.60±0.27 & 69.87±0.49 & 57.09±0.56 & 76.78  \\
    AUM+CT & \textbf{93.72±0.14} & \textbf{92.61±0.22} & \textbf{80.23±4.81} & \textbf{73.98±0.20} & \textbf{70.60±0.43} & \textbf{59.05±0.65} & 78.37  \\
    \midrule
    DIST  & 93.34±0.19 & 91.85±0.35 & 78.59±5.87 & 73.71±0.34 & 70.72±0.37 & 62.55±0.97 & 78.46  \\
    DIST+CT & \textbf{93.70±0.11} & \textbf{92.29±0.28} & \textbf{81.46±4.58} & 73.85±0.20 & \textbf{71.20±0.47} & 62.80±0.53 & 79.22  \\
    \bottomrule
    \end{tabular}%
  \caption{Comparison with SOTA sample selection methods in test accuracy (\%) on CIFAR-10 and CIFAR-100 with synthetic instance-dependent noise. The results are averaged over five random seeds. We use the Wilcoxon signed-rank test with a confidence level of 0.05 to compare the performance and bold the results where CT brings significant improvements.}
  \label{tab:IDN}
\end{table*}

\begin{figure*}[t]
    \includegraphics[width=\textwidth]{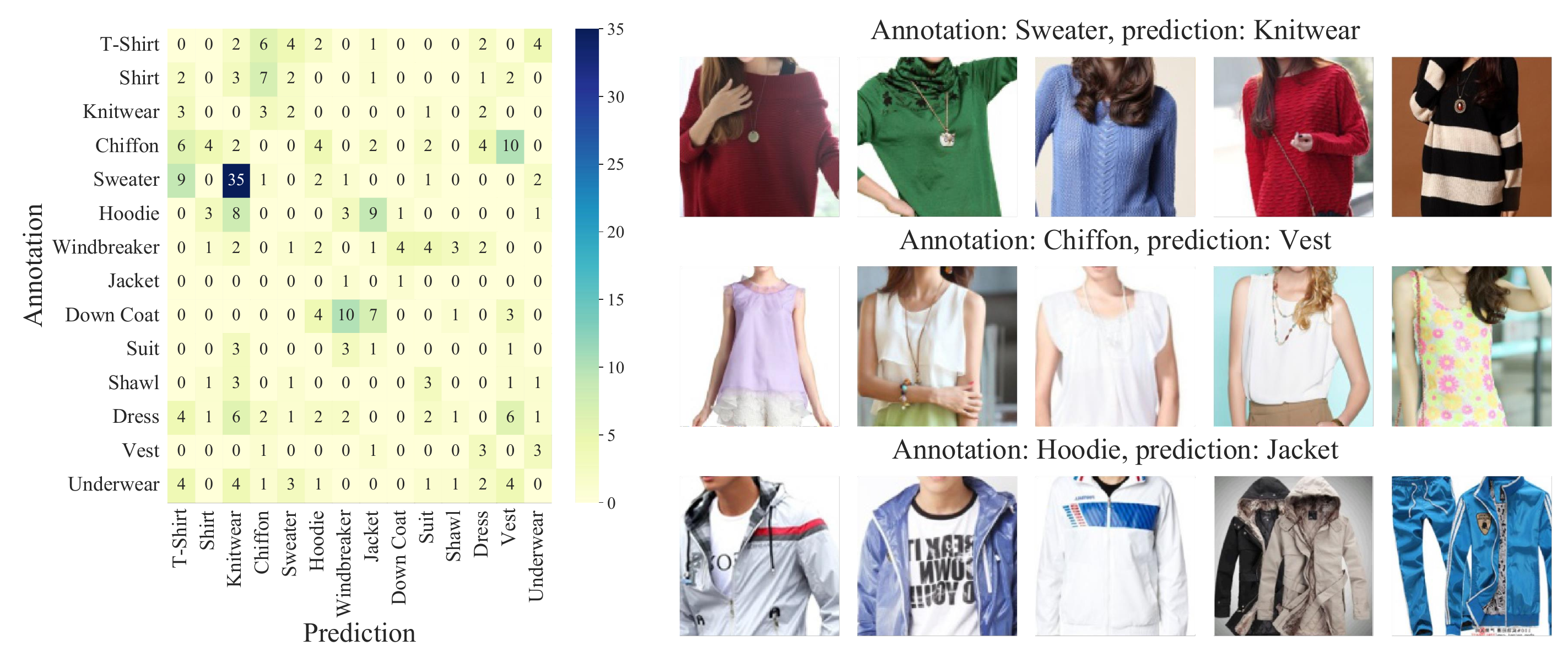}
    \caption{Left graph presents the number of test samples in which DivideMix predictions are consistent with the annotations while DivideMix+CT gives different predictions. The right graph shows cases.}
    \label{fig:clothing1m_badcase}
\end{figure*}

\begin{figure*}[t!]
    \centering
    \includegraphics[width=\textwidth]{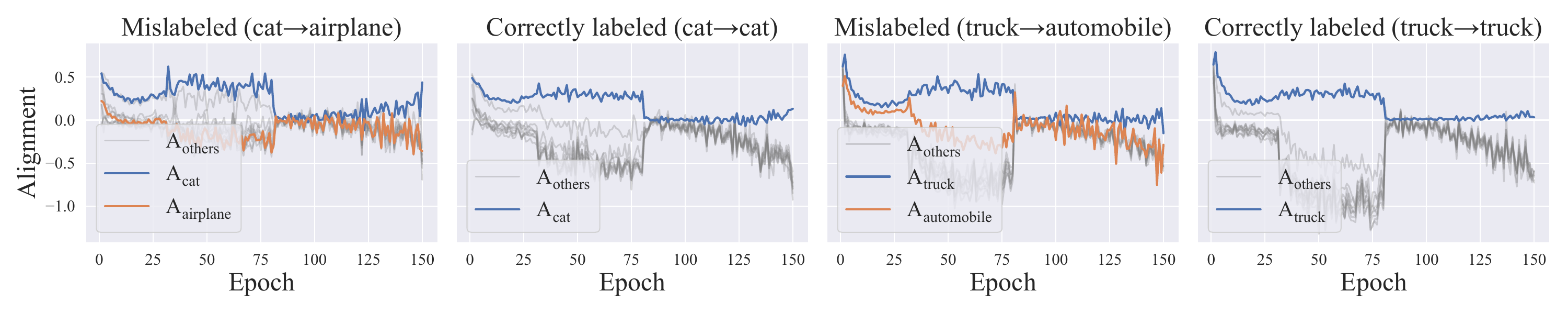}
    \caption{Gradient alignment when training a PreActResNet-18 net on CIFAR-10N-worst noisy labels when using GMM+CT as sample selector.}
    \label{fig:alignment_with_GMMCT}
\end{figure*}

\begin{figure*}[t!]
    \centering
    \includegraphics[width=\linewidth]{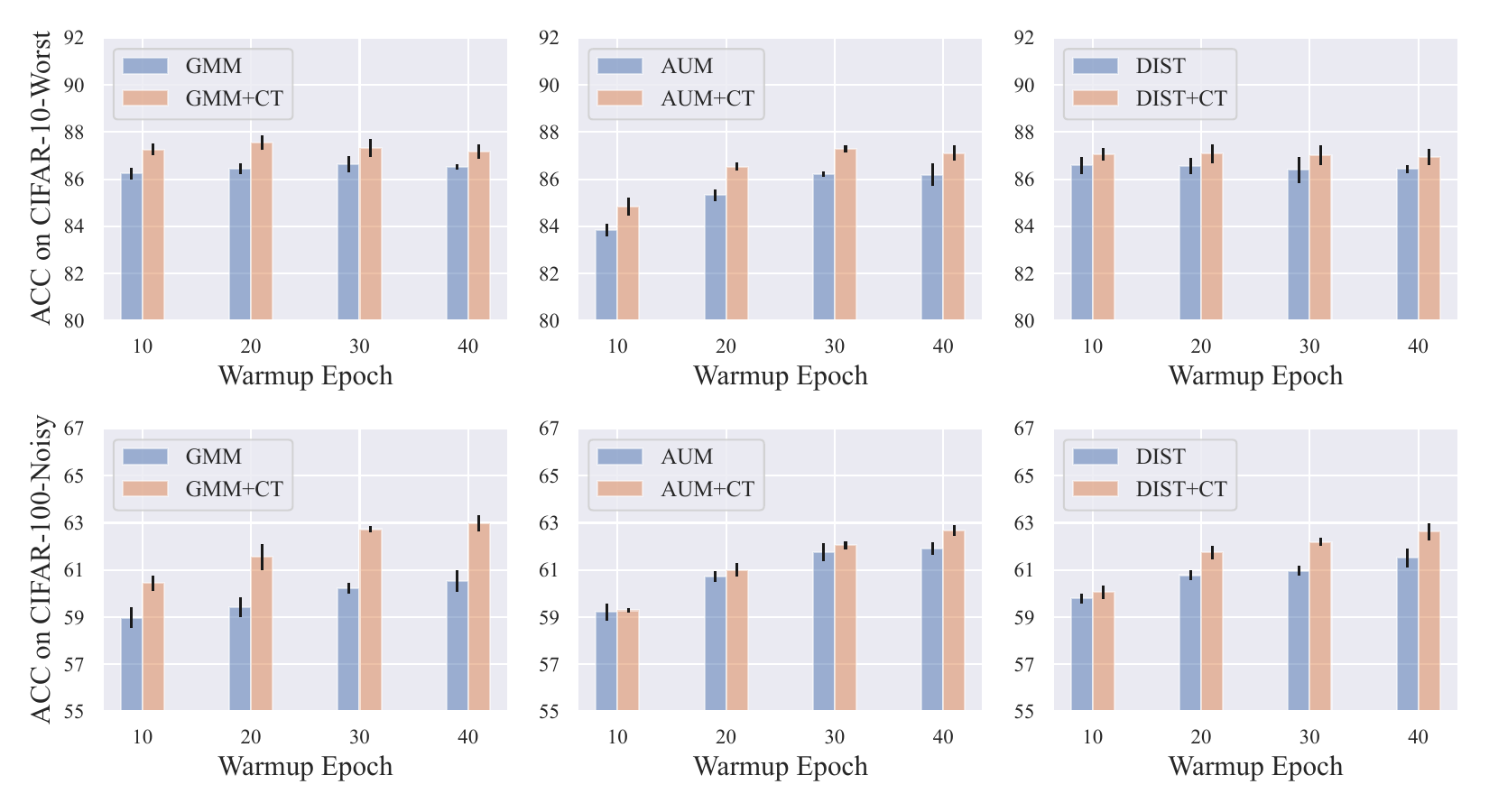}
    \caption{Test accuracy (\%) of different methods on CIFAR-10-Worst and CIFAR-100-Noisy with different warm-up epochs.}
    \label{fig:parameter_sensitivity_warmup}
\end{figure*}

\section{Additional Results}\label{appendix:additional_results}
Table~\ref{tab:cifarn} provides the complete results on the CIFAR-N dataset, which shows CT consistently improves existing sample selection methods on real-world noisy datasets. Table~\ref{tab:IDN} reports the performance of GMM, AUM, and DIST (the three best-performing sample selection methods in our experiments) under synthetic instance-dependent noise. We obtained instance-dependent noisy labels of CIFAR-10 and CIFAR-100 datasets according to \citeauthor{PDTM} and set the noise rate at values of 0.1, 0.3, and 0.5. We can see CT brings consistent improvement to all three baselines.

We also conducted experiments on the Clothing1M dataset~\cite{clothing1m}, where DivideMix achieved an accuracy of 74.12\% and DivideMix+CT achieved an accuracy of 74.18\%. However, we found the test set of Clothin1M contains noise labels. 
Figure~\ref{fig:clothing1m_badcase} shows the samples in which DivideMix predictions are consistent with the annotations while DivideMix+CT gives different predictions. We can see both the prediction of DivideMix+CT and annotations are correct for some test images since some classes have semantic overlaps like \quotes{Chiffon} and \quotes{Vest}, \quotes{Sweater} and \quotes{Knitwear}. Also, some labels are incorrect. The 2nd, 3rd, and 5th pictures in the last row on the right side of Figure~\ref{fig:clothing1m_badcase} should all be \quotes{Jacket} instead of \quotes{Hoodie}.
Recent study~\cite{DISC} also reports the test set of Clothin1M contains noise labels. Therefore, we do not use Clothing1M as the benchmark dataset.

\begin{table*}[t!]
  \centering
  \setlength{\tabcolsep}{1mm}
    \begin{tabular}{llcccccccc}
    \toprule
    \multicolumn{2}{l}{Dataset} & \multicolumn{4}{c}{CIFAR-10}  & \multicolumn{4}{c}{CIFAR-100} \\
    \midrule
    \multicolumn{2}{l}{Noise type} & Sym. 20\% & Sym. 50\% & Asym. 40\% & Real. 40\% & Sym. 20\% & Sym. 50\% & Asym. 40\% & Real. 40\% \\
    \midrule
    \multicolumn{2}{l}{GMM} & 91.60±0.28 & 88.07±0.08 & 89.45±0.85 & 86.63±0.34 & 69.59±0.32 & 63.95±0.44 & 65.29±0.42 & 60.22±0.23 \\
    \midrule
    \multirow{3}[1]{*}{GMM+CT} & $\alpha$=0.01 & 92.57±0.12 & 89.11±0.21 & 90.55±0.22 & 87.33±0.38 & 71.37±0.67 & 65.17±0.39 & 68.84±0.47 & 62.73±0.14 \\
          & $\alpha$=0.05 & 92.72±0.18 & 88.92±0.23 & 90.89±0.29 & 86.97±0.38 & 72.13±0.38 & 65.06±0.50 & 67.86±0.20 & 62.66±0.44 \\
          & $\alpha$=0.10 & 92.75±0.24 & 88.99±0.13 & 90.96±0.60 & 86.49±0.55 & 72.32±0.29 & 65.35±0.27 & 67.16±0.50 & 63.09±0.37 \\
    \midrule
    \multicolumn{2}{l}{FINE} & 89.13±0.48 & 85.66±0.32 & 82.56±2.00 & 80.09±0.45 & 70.96±0.45 & 58.58±0.48 & 49.48±0.77 & 56.87±0.25 \\
    \midrule
    \multirow{3}[1]{*}{FINE+CT} & $\alpha$=0.01 & 92.48±0.21 & 87.56±0.13 & 86.92±0.51 & 84.22±0.49 & 71.17±0.32 & 58.76±0.41 & 53.16±0.88 & 58.52±0.15 \\
          & $\alpha$=0.05 & 92.69±0.19 & 87.25±0.30 & 87.33±0.74 & 84.30±0.44 & 71.31±0.41 & 58.89±0.42 & 54.27±0.63 & 58.04±0.55 \\
          & $\alpha$=0.10 & 92.78±0.18 & 87.45±0.69 & 87.71±0.74 & 84.38±0.81 & 71.22±0.21 & 58.74±0.44 & 53.70±0.92 & 58.53±0.59 \\
    \midrule
    \multicolumn{2}{l}{AUM} & 92.31±0.13 & 87.80±0.24 & 88.21±0.54 & 86.22±0.11 & 72.50±0.44 & 64.90±0.28 & 61.25±0.41 & 61.75±0.38 \\
    \midrule
    \multirow{3}[1]{*}{AUM+CT} & $\alpha$=0.01 & 92.45±0.13 & 87.91±0.40 & 89.70±0.40 & 87.29±0.16 & 72.56±0.18 & 64.99±0.41 & 63.80±0.37 & 62.05±0.18 \\
          & $\alpha$=0.05 & 92.57±0.15 & 87.82±0.30 & 90.50±0.29 & 87.25±0.28 & 72.43±0.12 & 64.91±0.53 & 65.00±0.24 & 62.23±0.23 \\
          & $\alpha$=0.10 & 92.70±0.29 & 88.08±0.33 & 90.70±0.22 & 87.19±0.20 & 72.66±0.17 & 64.97±0.40 & 65.54±0.16 & 62.66±0.17 \\
    \midrule
    \multicolumn{2}{l}{DIST} & 92.63±0.15 & 88.43±0.24 & 90.00±0.42 & 86.39±0.54 & 72.73±0.32 & 65.59±0.24 & 66.74±0.81 & 60.97±0.20 \\
    \midrule
    \multirow{3}[1]{*}{DIST+CT} & $\alpha$=0.01 & 92.43±0.31 & 88.35±0.16 & 90.58±0.21 & 87.01±0.43 & 72.78±0.27 & 65.51±0.16 & 69.05±0.47 & 62.18±0.17 \\
          & $\alpha$=0.05 & 92.62±0.18 & 88.43±0.26 & 90.68±0.35 & 87.13±0.27 & 72.56±0.22 & 65.39±0.13 & 68.65±0.46 & 62.48±0.13 \\
          & $\alpha$=0.10 & 92.61±0.39 & 88.36±0.33 & 90.70±0.56 & 86.81±0.70 & 72.52±0.05 & 65.43±0.49 & 67.82±0.29 & 62.46±0.35 \\
    \bottomrule
    \end{tabular}%
   \caption{The test accuracy (\%) under different label noise when using different significance levels. The results are averaged over five random seeds.}
  \label{tab:sensitive}
\end{table*}

\section{Robustness Analysis}\label{appendix:robustness}
Recall that in Figure~\ref{fig:gradient_alignment_CE}, mislabeled samples eventually dominate the optimization process and the model is more encouraged to overfit wrong labels. CT would select incorrect labels in such a situation. However, these results are obtained without sample selection. Figure~\ref{fig:alignment_with_GMMCT} shows gradient alignment when using GMM+CT. After using proper sample selection methods to remove the mislabeled samples, the gradient alignment with respect to the noisy labels is not the highest during most of the training process which means CT will not select those mislabeled samples, ensuring the reliability of CT.

Most sample selection methods including small-loss criterion and our method all rely on the early learning phenomenon~\cite{memorizationeffect, ELR}, \ie model will fit correct labels first, before fitting mislabeled samples. Therefore, the number of warm-up epochs before performing sample selection may affect model performance. Too few warm-up epochs lead to underfitting to correct labels, while too many warm-up rounds lead to overfitting of mislabeled samples, both of which will harm sample selection. Fortunately, the results in Figure~\ref{fig:parameter_sensitivity_warmup} demonstrate both small-loss criterion and our method is not sensitive to the choice of warm-up epochs.

Additionally, integrating CT into existing sample selection methods only adds one hyperparameter, the significance level $\alpha$. Table~\ref{tab:sensitive} shows the CT is not sensitive to the choice of $\alpha$. This underscores the practical applicability and ease of implementation of the CT-enhanced sample selection approach in diverse experimental settings.

\begin{table*}[htbp]
  \centering
  \setlength{\tabcolsep}{1mm}
    \begin{tabular}{lcccccccc}
    \toprule
    Dataset & \multicolumn{4}{c}{CIFAR-10}  & \multicolumn{4}{c}{CIFAR-100} \\
    \midrule
    Noise type & Sym. 20\% & Sym. 50\% & Asym. 40\% & Real. 40\% & Sym. 20\% & Sym. 50\% & Asym. 40\% & Real. 40\% \\
    \midrule
    Original AUM & 92.23±0.21 & 85.18±0.52 & 83.00±1.04 & 80.66±0.81 & 72.01±0.13 & 63.19±0.64 & 55.26±0.75 & 60.59±0.79 \\
    Our implementation & 92.31±0.13 & 87.80±0.24 & 88.21±0.54 & 86.22±0.11 & 72.50±0.44 & 64.90±0.28 & 61.25±0.41 & 61.75±0.38 \\
    \bottomrule
    \end{tabular}%
  \caption{Test accuracy (\%) of different AUM implementations under different label noise. The results are averaged over five random seeds.}
  \label{tab:AUM}
\end{table*}

\section{Training Details}\label{appendix:train_details}
\subsection{Implementation Details}
\paragraph{L2D~\cite{L2D}.} 
This method trains LSTM networks in a supervised manner as noise detectors. We follow the author's instructions and train noise detectors on CIFAR-10 with 20\% symmetric label noise and CIFAR-100 with 30\% label noise. The former is used to detect wrong labels on CIFAR-10, and the latter is used for CIFAR-100.

\paragraph{Co-teaching~\cite{Co-teaching}.} 
The method consists of two networks with the same architecture but different weight initialization. They select the correct labels for each other during training. For each mini-batch, Co-teaching sorts samples according to per-sample loss and selects the bottom $R(T)\%$ samples for training. Here $R(T)\%$ is the sample selection schedule, which controls how many samples are selected in iteration $T$. Follows the original implementation, $R(T)=1-\min\{\frac{T}{T_{warmup}}\epsilon, \epsilon\}$, where $\epsilon$ is the estimated noise rate and $T_{warmup}$ is the number of epochs for warm-up. For a fair comparison, we only use one network for the test rather than integrate predictions from both networks.

\paragraph{CNLCU~\cite{CNLCU}.}
This method uses the lower confidence bounds of the loss values rather than the loss values themselves to select the correct labels. Its implementations are similar to Co-teaching.

\paragraph{GMM~\cite{DividMix}.} 
We collect per-sample loss on the fly during training and perform sample selection in each iteration after a period of warm-up. Before fitting a two-component Gaussian mixture model~(GMM), we use max-min normalization to normalize the per-sample loss to $[0,1]$. The hyperparameters in GMM are empirically set as: $\{max\_iter: 10, tol: 1e-2, reg\_covar: 1e-6\}$. GMM considers the probability of each sample belonging to the Gaussian distribution with a lower mean value as the correct probability of its label. Labels with correct probability higher than $\tau$ are selected for training. Here, $\tau$ is the global threshold, we choose it from $\{0.5, 0.7, 0.9, 0.95, 0.99\}$ according to the performance on the validation sets.

\paragraph{FINE~\cite{FINE}.}
This method first gets the principal components of latent representations for each class by eigendecomposition. Then FINE computes the similarity between sample representation and the principal component of its annotated class. Finally, a Gaussian mixture model is used to fit those similarity scores. FINE considers the probability of each sample belonging to the Gaussian distribution with a higher mean value as the correct probability of its label. Following the original implementation, the hyperparameters in GMM are set as $\{max\_iter:10, tol:1e-3, reg\_covar:1e-6\}$ and labels with correct probabilities higher than 0.5 are selected for training.

\paragraph{AUM~\cite{AUM}.}
The original AUM method trains the network until the first learning rate decay then calculates the average pre-logit margin for each sample to conduct sample selection. This implementation only performs sample selection once, while recent studies typically select correct labels in each iteration, which usually leads to a better performance. To this end, we follow the progressive sample selection strategy in our AUM implementation. Specifically, we collect model predictions on the fly and sort samples according to their average pre-logit margins over previous iterations. After a period of warm-up, we select the top $1-noise\_rate-k\%$ samples for training in each iteration. $k$ is a hyper-parameter that balances the sample selection precision and recall. We choose $k$ from $\{0.00, 0.05, 0.10\}$ according to the performance on the validation sets. Note we do not take the original \emph{threhsold strategy} since it would generate inappropriate thresholds for asymmetric label noise just as discussed in the original paper of AUM~\cite{AUM}. Table~\ref{tab:AUM} compares the performance of the original AUM and our implementation. We can see our implementation significantly improves the performance.

\paragraph{DIST~\cite{DISC}.}
This method maintains a dynamic threshold $\tau_i$ for the $i$th training sample, which is updated by: $\tau_i = \lambda \tau_i + (1 - \lambda) \max{p_i}$. Here $\lambda$ is the momentum hyperparameter and $\max{p_i}$ is the maximum confidence of the model prediction for the $i$th training sample. DIST selects samples when the model's prediction confidence for the annotated labels is above thresholds. We choose $\lambda$ from $\{0.90, 0.95, 0.99\}$ according to the performance on the validation sets. Similar to our other implementations, we collect model prediction on the fly to update $\tau_i$ and perform sample selection in each iteration after a period of warm-up.

\paragraph{HMW~\cite{HMW}.}
This method use average margins in logits to reweight every training example for better performance. We strictly follow the original implementation, please refer to the original paper or our code for more details.  

\paragraph{Advanced LNL methods.}
For advanced LNL methods in Table~\ref{tab:SOTA+CT}, we strictly follow the implementation of the public code, please refer to their original papers or our code for more implementation details.

\begin{table*}[t!]
  \centering
  \setlength{\tabcolsep}{1mm}
    \begin{tabular}{lccccccc}
    \toprule
    \multicolumn{2}{c}{\multirow{2}[4]{*}{Hyperparameter}} & \multicolumn{3}{c}{CIFAR-10} & \multicolumn{3}{c}{CIFAR-100} \\
\cmidrule{3-8}    \multicolumn{2}{c}{} & Sym. 20\% & Sym. 50\% & Asym. 40\% & Sym. 20\% & Sym. 50\% & Asym. 40\% \\
    \midrule
    GMM   & $\tau$   & 0.50  & 0.50  & 0.50  & 0.50  & 0.50  & 0.90  \\
    GMM+CT & $\tau$   & 0.50  & 0.50  & 0.90  & 0.50  & 0.50  & 0.90  \\
    \midrule
    FINE  & $\tau$   & 0.50  & 0.50  & 0.50  & 0.50  & 0.50  & 0.50  \\
    FINE+CT & $\tau$   & 0.50  & 0.50  & 0.50  & 0.50  & 0.50  & 0.50  \\
    \midrule
    DIST  & $\lambda$ & 0.99  & 0.99  & 0.95  & 0.99  & 0.99  & 0.95  \\
    DIST+CT & $\lambda$ & 0.99  & 0.99  & 0.90  & 0.99  & 0.99  & 0.90  \\
    \midrule
    AUM   & $k$     & 0.05  & 0.00  & 0.05  & 0.05  & 0.05  & 0.10  \\
    AUM+CT & $k$     & 0.05  & 0.00  & 0.05  & 0.05  & 0.05  & 0.10  \\
    \bottomrule
    \end{tabular}%
  \caption{Hyperparameter configuration on CIFAR-10 and CIFAR-100 datasets with symmetric and asymmetric label noise.}
  \label{tab:hyperparameter_cifar_sym_asym}%
\end{table*}%

\begin{table*}[t!]
  \centering
  \setlength{\tabcolsep}{1mm}
    \begin{tabular}{lccccccc}
    \toprule
    \multicolumn{2}{c}{\multirow{2}[4]{*}{Hyperparameter}} & \multicolumn{3}{c}{CIFAR-10} & \multicolumn{3}{c}{CIFAR-100} \\
\cmidrule{3-8}    \multicolumn{2}{c}{} & IDN. 10\% & IDN. 30\% & IDN. 50\% & IDN. 10\% & IDN. 30\% & IDN. 50\% \\
    \midrule
    GMM   & $\tau$   & 0.50  & 0.70  & 0.90  & 0.50  & 0.50  & 0.90  \\
    GMM+CT & $\tau$   & 0.95  & 0.95  & 0.99  & 0.99  & 0.70  & 0.95  \\
    \midrule
    DIST  & $\lambda$ & 0.99  & 0.99  & 0.95  & 0.99  & 0.99  & 0.95  \\
    DIST+CT & $\lambda$ & 0.90  & 0.95  & 0.90  & 0.90  & 0.95  & 0.95  \\
    \midrule
    AUM   & $k$     & 0.05  & 0.05  & 0.10  & 0.10  & 0.10  & 0.10  \\
    AUM+CT & $k$     & 0.10  & 0.10  & 0.10  & 0.10  & 0.10  & 0.10  \\
    \bottomrule
    \end{tabular}%
  \caption{Hyperparameter configuration on CIFAR-10 and CIFAR-100 datasets with synthesized instance-dependent label noise.}
  \label{tab:hyperparameter_cifar_IDN}
\end{table*}

\begin{table*}[t!]
  \centering
  \setlength{\tabcolsep}{1mm}
    \begin{tabular}{lccccccccc}
    \toprule
    \multicolumn{2}{c}{\multirow{2}[4]{*}{Hyperparameter}} & \multicolumn{6}{c}{CIFAR-10N}                 & \multicolumn{2}{c}{CIFAR-100N} \\
\cmidrule{3-10}    \multicolumn{2}{c}{} & Clean & Aggregate & Random1 & Random2 & Random3 & Worst & Clean & Noisy \\
    \midrule
    GMM   & $\tau$   & 0.50  & 0.50  & 0.50  & 0.50  & 0.50  & 0.50  & 0.50  & 0.50  \\
    GMM+CT & $\tau$   & 0.50  & 0.50  & 0.50  & 0.50  & 0.50  & 0.50  & 0.50  & 0.50  \\
    \midrule
    FINE  & $\tau$   & 0.50  & 0.50  & 0.50  & 0.50  & 0.50  & 0.50  & 0.50  & 0.50  \\
    FINE+CT & $\tau$   & 0.50  & 0.50  & 0.50  & 0.50  & 0.50  & 0.50  & 0.50  & 0.50  \\
    \midrule
    DIST  & $\lambda$ & 0.99  & 0.95  & 0.95  & 0.95  & 0.95  & 0.95  & 0.99  & 0.95  \\
    DIST+CT & $\lambda$ & 0.99  & 0.95  & 0.95  & 0.95  & 0.95  & 0.90  & 0.99  & 0.90  \\
    \midrule
    AUM   & $k$     & 0.05  & 0.05  & 0.05  & 0.05  & 0.05  & 0.05  & 0.05  & 0.00  \\
    AUM+CT & $k$     & 0.05  & 0.05  & 0.05  & 0.05  & 0.05  & 0.05  & 0.05  & 0.00  \\
    \bottomrule
    \end{tabular}
  \caption{Hyperparameter configuration on CIFARN datasets.}
  \label{tab:hyperparameter_cifarn}
\end{table*}

\begin{table*}[htbp]
  \centering
  \setlength{\tabcolsep}{1mm}
    \begin{tabular}{lcccccccc}
    \toprule
    \multicolumn{1}{c}{Method} & batch size & optimizer &learning rate    & momentum & weight decay    & epoch & warm-up & milestone \\
    \midrule
    \multicolumn{9}{c}{miniWebVision} \\
    \midrule
    DivideMix & 32  & SGD  & 0.01  & 0.9   & 5e-4 & 100   & 1     & 50 \\
    F-DivideMix & 32  & SGD  & 0.01  & 0.9   & 5e-4 & 100   & 1     & 50 \\
    CORES & 64  & SGD  & 0.01  & 0.9   & 5e-4 & 100   & 10    & 50 \\
    DISC  & 32  & SGD  & 0.20   & 0.0     & 5e-4 & 100   & 15    & 50, 80 \\
    \midrule
    \multicolumn{9}{c}{Food-101N} \\
    \midrule
    DivideMix & 32  & SGD  & 0.002 & 0.9   & 5e-4 & 30    & 1     & 10, 20 \\
    F-DivideMix & 32  & SGD  & 0.002 & 0.9   & 5e-4 & 30    & 1     & 10, 20 \\
    CORES & 64 & SGD   & 0.002 & 0.9   & 5e-4 & 30    & 1     & 10, 20 \\
    DISC  & 32  & SGD  & 0.010  & 0.0     & 5e-4 & 30    & 5     & 10, 20 \\
    \bottomrule
    \end{tabular}%
  \caption{Hyperparameter configuration on real-world noisy datasets.}
  \label{tab:hyperparameter_real}%
\end{table*}

\subsection{Hyperparameter Configuration}
Next, we introduce the training process and hyperparameter configuration. For all the experiments on CIFAR-10/CIFAR-100/CIFAR-N datasets, we use PreActResNet-18~\cite{ResNet, PreActResNet} as the backbone and train it for 150 epochs using SGD with a momentum of 0.9, a weight decay of 0.001, and a batch size of 128. The initial learning rate is set as 0.02 and reduced to 0.002 after 80 epochs. The number of warm-up epochs is 30.
Following~\cite{ELR}, we retain 10\% of the training sets to perform validation and determine the hyperparameter configuration by performance on the validation set. Table~\ref{tab:hyperparameter_cifar_sym_asym}, Table~\ref{tab:hyperparameter_cifar_IDN} and Table~\ref{tab:hyperparameter_cifarn} provide hyperparameter configurations for each method.

Table~\ref{tab:hyperparameter_real} provides the hyperparameter configurations for CORES~\cite{CORES}, DivideMix~\cite{DividMix}, f-DivideMix~\cite{FINE}, and DISC~\cite{DISC} on real-world noisy datasets. When integrating CT with those advanced LNL methods, the significance level $\alpha$ is empirically set to $0.10$ without further tuning, and other hyperparameters are the same with Table~\ref{tab:hyperparameter_real}.

\begin{table*}[t!]
  \setlength{\tabcolsep}{1mm}
  \centering
    \begin{tabular}{cccc}
    \toprule
    Dataset & DISC  & FINE  & GMM \\
    \midrule
    miniWebVision & $\lambda$=0.99 & max\_iter=10, tol=1e-6, reg\_covar=1e-6, $\tau$=0.5 & max\_iter=10, tol=1e-2, reg\_covar=5e-4, $\tau$=0.5 \\
    Food101N & $\lambda$=0.97 & max\_iter=10, tol=1e-6, reg\_covar=1e-6, $\tau$=0.5 & max\_iter=10, tol=1e-2, reg\_covar=5e-4, $\tau$=0.5 \\
    \bottomrule
    \end{tabular}
  \caption{The hyperparameter configurations used when performing linear probing with sample selection using CLIP features.}
  \label{tab:hyperparameter_linear_prob}%
\end{table*}

\begin{table}[t!]
  \centering
  \setlength{\tabcolsep}{1mm}
    \begin{tabular}{lccc}
    \toprule
    \multicolumn{1}{c}{Method} & Webvision & ILSCRC12 & Food-101N \\
    \midrule
    no sample selection & 86.39±0.08 & 84.89±0.11 & 93.33±0.02 \\
    \midrule
    GMM   & 86.40±0.15 & 85.62±0.18 & 94.06±0.02 \\
    GMM+CT & \textbf{86.56±0.14} & \textbf{85.90±0.19} & 94.05±0.03 \\
    \midrule
    FINE  & 85.84±0.14 & 85.06±0.12 & 93.25±0.02 \\
    FINE+CT & 85.97±0.09 & 85.02±0.11 & \textbf{93.38±0.02} \\
    \midrule
    DIST  & 86.58±0.09 & 85.51±0.05 & 93.95±0.02 \\
    DIST+CT & \textbf{86.74±0.11} & \textbf{85.62±0.09} & \textbf{94.02±0.02} \\
    \bottomrule
    \end{tabular}
  \caption{Test accuracy (\%) when performing linear probing using the CLIP feature with different sample selection methods. The results are averaged over 10 random seeds. We use a t-test with a confidence level of 0.05 to compare the performance and bold the results with significant improvements.}
  \label{tab:linear_prob}
\end{table}

\section{Further Discussion}\label{appendix:further_discussion}
Recent study~\cite{LRA-diffusion} introduces strong CLIP features to boost the performance. In this paper, we explore whether our method can benefit from CLIP features. To keep it simple, we directly conduct sample selection when performing linear probing using CLIP features (ViT-L/14). The whole training process is the same as Algorithm~\ref{alg:ct}, where the training sample $\vec{x}^{[i]}$ is the CLIP feature of the $i$th sample and the neural network $f(\cdot;\theta)$ is instanced by a linear layer. 
We add a dropout layer before the linear layer to enhance performance, the dropout rate is set to 0.3. We train the linear layer for 100 epochs using SGD with a momentum of 0.9, a weight decay of 0.001, and a batch size of 10240. The initial learning rate is set as 0.1 and reduced to 0.01 after 50 epochs. The significant level $\alpha$ used in CT is set to 0.5.
Table~\ref{tab:hyperparameter_linear_prob} gives other hyperparameter configurations.
Table~\ref{tab:linear_prob} shows that CT can still enhance existing sample selection methods when using CLIP features.

We further analyzed the computational efficiency of CT. All the experiments in this paper were taken on a system with an RTX4090 GPU and 128G RAM. We use mixed precision to speed up the training process. For sample selection methods, Co-teaching and CNLCU take around 40 minutes for one trail, and other methods take around 25 minutes. For advanced LNL methods, Table~\ref{tab:time} shows the time needed to conduct experiments on miniWebVision and Food-101N. Notably, CT is highly computationally efficient, enhancing algorithm performance without substantially increasing the total computation time.

\begin{table}[t!]
  \centering
    \begin{tabular}{lcc}
    \toprule
    \multicolumn{1}{c}{Method} & miniWebVision & Food-101N \\
    \midrule
    CORES & 4.38h & 5.26h \\
    CORES+CT & 4.40h & 5.14h \\
    \midrule
    DivideMix & 54.13h & 23.65h \\
    DivideMix+CT & 56.22h & 23.44h \\
    \midrule
    f-DivideMix & 60.44h & 34.92h \\
    f-DivideMix+CT & 62.50h & 36.74h \\
    \midrule
    DISC  & 18.05h & 8.77h \\
    DISC+CT & 18.38h & 8.09h \\
    \bottomrule
    \end{tabular}%
  \caption{Total training time in hours on miniWebVision and Food-101N.}
  \label{tab:time}%
\end{table}%

\section{Limitations and Broader Impacts}\label{appendix:limitations}
The main limitation of CT is the need to monitor model predictions for each sample. Usually, CT only needs to collect model predictions on the fly. However, on some datasets, the model may be able to fit noisy labels within the first few epochs. Under such a situation, CT can no longer collect the model’s predictions on the fly. A feasible method is to sample a part of the data for training in each iteration and collect the model predictions for every example after each iteration. This incurs additional computational overhead. But it’s worth noting that collecting model predictions and training the model can be performed in parallel. For example, a copy of the model at the end of each epoch can be saved to collect model predictions while continuing to train the model. Such parallel processing can reduce additional training time.

In terms of positive social impact, LNL techniques can handle imperfect labels, which makes the development of machine learning models more efficient as it reduces the need for labor-intensive perfect labeling. In terms of negative social impact, if LNL is used in technologies that make decisions impacting specific groups like recommendation systems, it could potentially lead to unfair outcomes. For example, the sample selection methods may generate biased results, treating some uncommon items as noisy labels. Using user feedback to verify the sample selection results is one possible solution.

\end{document}